\documentclass[sigconf,preprint]{acmart}
\usepackage{adjustbox}
\usepackage{multirow}
\usepackage{subcaption}
\usepackage{stfloats}
\usepackage{algorithm}
\usepackage{algorithmic}
\AtBeginDocument{%
  }

\renewcommand\footnotetextcopyrightpermission[1]{}

\begin{document}

\title{ViSMaP: Unsupervised Hour-long Video Summarisation by Meta-Prompting}


\author{
Jian Hu$^{1,2}$\textsuperscript{*} \quad
Dimitrios Korkinof$^{2}$ \quad
Shaogang Gong$^{1}$ \quad
Mariano Beguerisse-Díaz$^{2}$\textsuperscript{\dag} \\
\textit{$^1$Queen Mary University of London \quad $^2$Spotify} \\
\text{\small jian.hu@qmul.ac.uk, dkorkinof@spotify.com, s.gong@qmul.ac.uk, marianob@spotify.com}
}





\renewcommand{\shortauthors}{Jian et al.}

\begin{abstract}
We introduce ViSMap: Unsupervised \textbf{Vi}deo \textbf{S}ummarisation by \textbf{M}et\textbf{a} \textbf{P}rompting, a system to summarise hour long videos with no-supervision.
Most existing video understanding models work well on short videos of pre-segmented events, yet they struggle to summarise longer videos where relevant events are sparsely distributed and not pre-segmented. Moreover, long-form video understanding often relies on supervised hierarchical training that needs extensive annotations which are costly, slow and prone to inconsistency.
With ViSMaP we bridge the gap between short videos (where annotated data is plentiful) and long ones (where it's not). We rely on LLMs to create optimised pseudo-summaries of long videos using segment descriptions from short ones. These pseudo-summaries are used as training data for a model that generates long-form video summaries, bypassing the need for expensive annotations of long videos.
Specifically, we adopt a meta-prompting strategy to iteratively generate and refine creating pseudo-summaries of long videos.
The strategy leverages short clip descriptions obtained from a supervised short video model to guide the summary.
Each iteration uses three LLMs working in sequence: one to generate the pseudo-summary from clip descriptions, another to evaluate it, and a third to optimise the prompt of the generator. This iteration is necessary because the quality of the pseudo-summaries is highly dependent on the generator prompt, and varies widely among videos.
We evaluate our summaries extensively on multiple datasets; our results show that ViSMaP achieves performance comparable to fully supervised state-of-the-art models while generalising across domains without sacrificing performance. 
Code will be released upon publication.
%
\end{abstract}


\maketitle

\renewcommand{\thefootnote}{\fnsymbol{footnote}}
\setcounter{footnote}{0}
\footnotetext[1]{Partly done during a Spotify internship.}
\footnotetext[2]{Corresponding author.}

\section{Introduction}

\begin{figure}[tp]
\centering
    \begin{subfigure}[t]{0.5\textwidth}
\includegraphics[width=1.0\columnwidth]{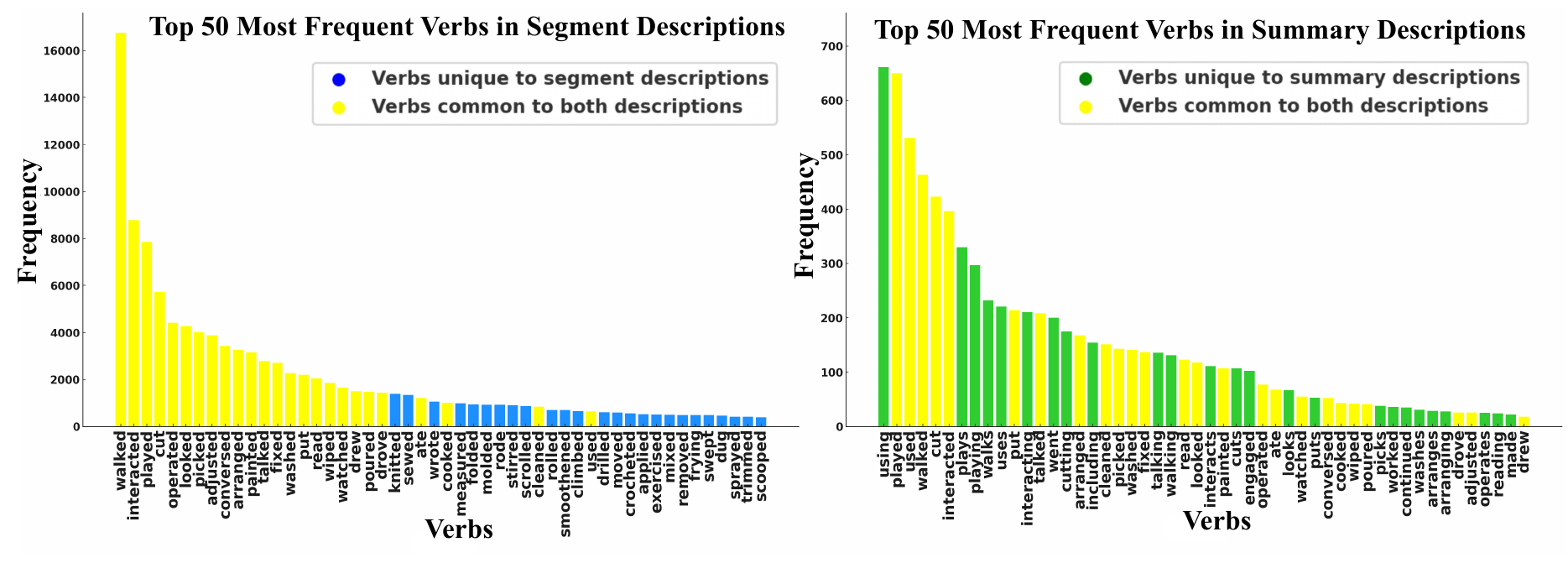}
 \vspace{-15pt}
        \caption{Verb frequency: Segments vs Summaries. Illustration of the semantic gap between segment descriptions and video summaries in Ego4D-HCap \cite{islam2024video} dataset.}
        \label{fig:verb_hist}
    \end{subfigure}
    \begin{subfigure}[t]{0.85\textwidth}
    \vspace{10pt}\hspace{-5pt}\includegraphics[width=0.6\columnwidth]{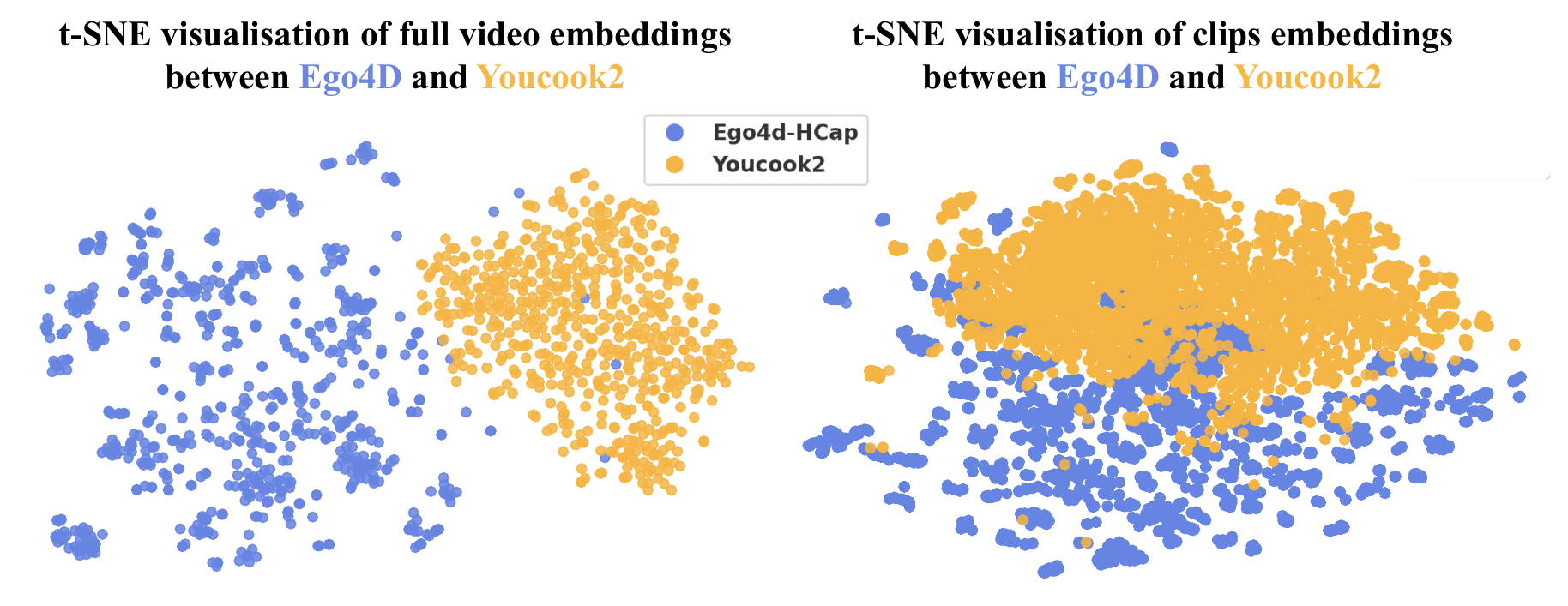}
        \caption{t-SNE visualization between Ego4D-HCap and Youcook2 datasets. \\Illustration of the distributional shift between these two dataset.}
        \label{fig:tsne}
    \end{subfigure}
\caption{Two main challenges we address with our approach: (a) Bridging the semantic gap between short-form segment descriptions and hour-long video summaries descriptions (b) Overcoming the domain shift between the source domain and the target domain.}
\end{figure}

Video captioning models are typically trained on large datasets composed of short videos along with their corresponding captions; these videos are usually shorter than three minutes~\cite{wang2018reconstruction, hori2017attention}. While this approach equips models to describe simple actions like walking or talking, it struggles with the complexity of long-form videos (videos longer than three minutes). These include videos, such as vlogs, sporting events, or movies, which are often complex and could be an hour or longer in duration. When used on hour-long videos, short-form video models are only able to produce descriptions for short segments of video that represent isolated actions without capturing the underlying {\em story}, resulting in extraneous details~\cite{islam2024video,lei2020mart}. For example, a model may capture individual actions in a biographical film, but it will struggle to summarise the significant events of a person's life.

\begin{figure*}[ht]
   \centering  
   \includegraphics[width=18cm, trim=0cm 0 0 0, clip]{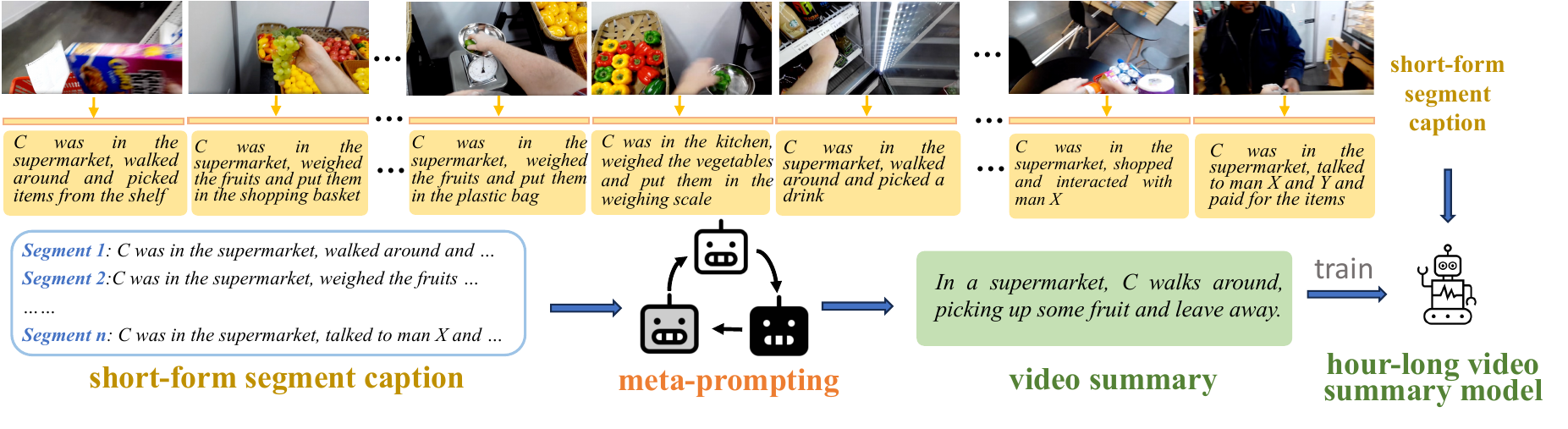} 
   \vspace{-15pt}
   \caption{ 
   Motivation of our VisMaP. Most existing video summarisation models focus on minute-level short-form videos, while hour-long videos, which are more common in real-world scenarios, are often overlooked due to their length, content complexity, and the prohibitively high cost of manual annotation. We propose a cross-domain unsupervised approach for hour-long video summarisation. It leverages the inductive power of multiple LLMs to generate high-quality pseudo-summaries from short video segments via meta-prompting. These pseudo-summaries are then used to train a model, enabling effective summarisation of long videos without costly human annotations.}
   \label{fig:motivation}
\end{figure*}

To extend the capabilities of video captioning to longer durations, MA-LMM \cite{he2024ma} and LaViLa~\cite{zhao2023learning} use LLMs to describe long-form videos with duration up to 10 minutes. These models still struggle with the extensive content of hour-long videos, for which there is a notable lack of suitable training datasets. 
Addressing this gap, Ego4D \cite{grauman2022ego4d} introduced the first extensive hour-long video dataset; however, these videos are mostly first-person, and they differ from the typical third-person videos (e.g., vlogs, documentaries, etc), which may limit their broader use in video understanding tasks.
To better summarise hour-long videos, the authors of Video Recap~\cite{islam2024video} introduced a recursive video captioning model trained on an hour-long video dataset with multi-granularity annotations. It identifies key moments from redundant information to generate effective summaries. However, annotating hour-long videos at multiple levels of granularity is not only costly but also prone to annotator inconsistencies~\cite{islam2024video}; this creates the need for careful quality control of the annotations, which imposes additional cost and scaling constraints. 
In contrast, extensive collections of annotated short-form videos are readily available for training models.

In this work, we propose a scalable approach to generate unsupervised hour-long video summaries. Our aim is to train a lightweight summarisation model from annotated short-form videos in a source domain, and then generalise to unannotated hour-long videos in a target domain. For this purpose, we develop methods to transition from recognising single actions in short videos and summarising the more complex activities in unannotated hour-long videos (See Fig.\ref{fig:motivation}). There are three main challenges in understanding complex behaviours in hour-long videos from their associated atomic actions. The first challenge is identifying the salient atomic actions within a large collection of redundant information. The second is bridging the semantic gap between atomic actions in short-form segments and their corresponding complex behaviour within long videos (see Fig.~\ref{fig:verb_hist}). A complex behaviour may consist of many atomic actions (where order may also matter) that collectively identify the behaviour, thus a certain level of reasoning is required to bridge the semantic gap between those atomic actions and the complex behaviour. Finally, we also address the semantic gap the may exist between the source and target domains due to differences in the content, as is the case for instance between the Ego4D-HCap and Youcook2 datasets (see Fig.~\ref{fig:tsne}).

\begin{figure*}[tp]
   \centering  
   \includegraphics[width=17.9cm]{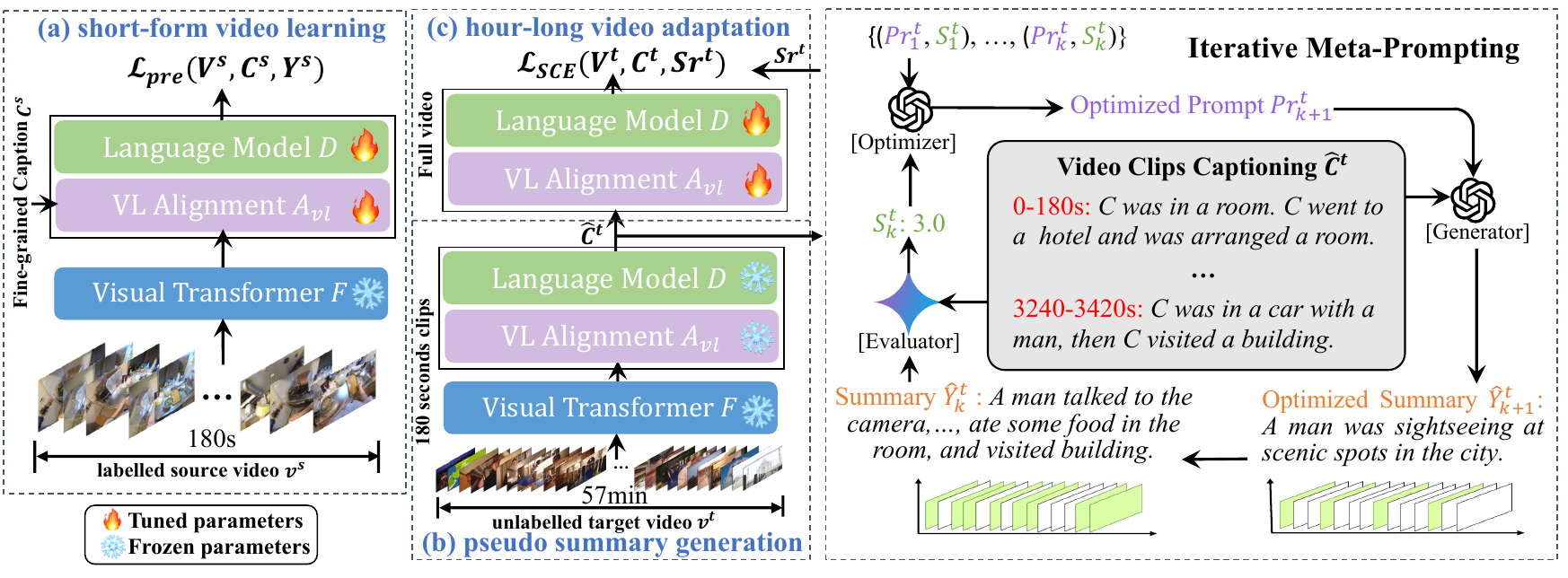}
   \vspace{-20pt}
   \caption{An overview of our VisMaP. \textbf{(a) First stage:} we use
     180-second source video $v^s$ for supervised pretraining to establish basic summary capabilities. \textbf{(b) Second stage:} we split hour-long target videos $v^t$
     into 3-minute segments set $\mathbf{V}_i^t$ and process them through the first-stage summary model to generate pseudo captions $\widehat{C}^t$. $\widehat{C}^t$ are then refined through a meta-prompting process with $K$ iterations, using Gemini as the evaluator and GPT-3.5
     as the optimiser and the generator, to create more tailored prompts ${Pr}^t$ and summaries $\widehat{Y}^t$. \textbf{(c) Third stage:} Refined $\widehat{Y}^t$ pseudo-summaries are utilised to fine-tune the summary model for effective hour-long video summary.
   }\label{fig:framework}
\end{figure*}

To address these challenges we introduce \textbf{Vi}deo \textbf{S}ummarisation by \textbf{M}et\textbf{a} \textbf{P}rompting (VisMaP). An overview of the framework is shown in Fig.~\ref{fig:framework}, and consists of three stages:
First, we use minute-long annotated videos as the source domain to construct a lightweight short-form video summary model. Second, we treat unsupervised hour-long videos as the target domain and split them into minute-long segments to reduce the semantic gap caused by video length. We then use the summary model from Stage 1 to generate pseudo-captions for these segments transfering the visual information into the text domain. After that, we use these pseudo-captions along with {\em meta-prompting} to construct optimised pseudo-summaries (i.e., summaries of long videos generated without human intervention) for hour-long target domain videos. Using the contextual reasoning and reflection capabilities of LLMs, our method iteratively discovers and selects relevant semantic tokens associated with atomic activities in the source domain and complex activities in the target domain, aiming to construct optimised pseudo-summaries.
Finally, the third challenge is to fine-tune the foundational model from Stage 1 to generate hour-long video summaries from minute-long segment pseudo-captions, using the optimised pseudo-summaries as supervision.
We evaluate the performance of VisMaP on five datasets, and we provide theoretical results showing that VisMaP reduces the error upper bound due to the domain gap caused by both video length and semantic distribution.
\textbf{Our contributions are the following:}
\begin{itemize}
    \item To our knowledge, we are the first to present and address the challenge of  unsupervised hour-long video summary.
    \item We formulate a mechanism to extend video understanding from annotated short-form videos to unlabelled hour-long ones. We do this by optimising semantic alignment, and bridging the gap caused by longer video lengths and semantic distribution. Specifically, we present meta-prompting, using LLMs to optimise the selection and localisation of key events in hour-long videos. With pseudo captions from models trained on short-form videos, our approach uses sequences of atomic action descriptions to iteratively obtain more accurate holistic summaries of hour-long videos.
    \item We conduct experiments on five datasets to demonstrate the effectiveness and robustness of ViSMaP. These experiments show that we can attain comparable performance to fully supervised state of the art models, in an unsupervised and, importantly, domain adaptable manner.
\end{itemize}

\vspace{-10pt}
\section{Related Works}

\noindent\textbf{Visual-Language Models} enhance cross-modality understanding and reasoning~\cite{zhang2024vision, bai2023qwen, wang2024qwen2}. CLIP~\cite{radford2021learning} and ALIGN \cite{jia2021scaling} were the first to expand language models to vision-language tasks; LLaVA~\cite{liu2024visual} and MiniGPT-4 \cite{zhu2023minigpt} combine LLM and image embeddings to give LLMs the ability to understand images. Subsequently, Refs.~\cite{li2023videochat, lin2023video} enable video understanding by modelling the temporal sequences of extracted video embeddings. Refs.~\cite{li2025llama,li2024llava} present further improvements on the video understanding capabilities by constructing higher-quality large-scale video datasets for training. In spite of their good results, one obstacle limiting their performance is the need for large long-form video datasets with extensive annotations, which are difficult and expensive to obtain. VisMaP sidesteps this challenge by requiring only short-form video annotations and adapting them to long-form videos. \\
\noindent\textbf{Short-form video models} typically process videos under 3 minutes and include tasks like: a) Video Question Answering~\cite{yang2022zero, li2022invariant}, which answers questions based on an understanding of the entire video clip; b) video captioning~\cite{wang2018reconstruction, iashin2020multi} goes further, requiring more detailed descriptions for different clips of a video; c) video grounding \cite{zeng2020dense, zhang2020does} locates the temporal moments in a video that correspond to a given text description. These tasks only require temporal or spatial localisation and understanding; hour-long video summarisation, on the other hand, is more complex because it needs to identify crucial frames within a vast amount of redundant information {\it and} to comprehend them. 

\noindent\textbf{Long-form Video Models} normally handle videos between 3 and 10 minutes; for example, Refs.~\cite{he2024ma, wang2024videollamb} optimise memory to retain context for longer video analysis; Refs.~\cite{li2025llama, wang2024longllava} filter visual tokens to expand the range of video lengths that can be processed;
LLoVi~\cite{zhang2023simple} uses LLMs to summarise short video clip captions. All of these models struggle with hour-long videos because identifying sporadically dispersed important moments among extensive information is difficult for them.
LongVA~\cite{zhang2024long} and LLaVA-Video~\cite{zhang2024video} are capable of performing VQA on hour-long videos, but they require large-scale annotated long video datasets for training and attain limited performance on video summarisation tasks. 
Video ReCap~\cite{islam2024video} uses recursive supervised training for hour-long first-person videos to identify important moments and create summaries, but requires extensive annotations and struggles with videos in third-person perspective.  VisMaP  avoids these problems by relying on meta-prompting to perform spatio-temporal modelling, and generate summaries from redundant information in long videos. \\
\noindent\textbf{Domain Adaptation} (DA) transfers useful information from a labelled source domain to an unlabelled target domain \cite{liu2022deep}. For example, conventional DA approaches require simultaneous access to both source and target domain data during training, and typically rely on adversarial learning~\cite{tzeng2017adversarial, hu2020discriminative} or high-order moment matching~\cite{Long16,inproceedings}. In contrast, {\it source-free} DA methods only need a model previously trained on the source domain for target adaptation ~\cite{liang2020we, kundu2020universal}; 
VisMaP falls into this category.  While previous work focus mostly on classification and segmentation, VisMaP is the first to our knowledge that applies DA methods to generate unsupervised hour-long video summaries. Furthermore, VisMaP not only identifies target atomic activities in source domains (with domain gaps), but also discovers undefined new complex activities.
\section{Methodology}
\subsection{Problem Definition}
We tackle a cross-domain video summarization task using a labelled short-form dataset $\mathcal{D}_s = {\{V_i^s, C_i^s, Y_i^s\}}_{i=1}^{n_s}$ as our source domain. Here, $V_i^s$ represents a 3-minute short-form video, $C_i^s$ denotes captions for each 4-second segment within the video, and $Y_i^s$ is the summary of the entire 3-minute video. In addition, we employ an unlabelled hour-long dataset $\mathcal{D}_t = {\{V_i^t\}}_{i=1}^{n_t}$ as our target domain. The variables $n_s$ and $n_t$ indicate the number of videos in the source and target domains, respectively. We make no assumptions about the distributions of $\mathcal{D}_s$ and $\mathcal{D}_t$; that is, data in the two domains can originate from different distributions  (e.g., the source domain could be first-person shopping videos, and the target domain is third-person cooking videos). After being trained on the fully-labelled source dataset $\mathcal{D}_s$, the model needs to generate a caption $Y_i^t = [y_{i,j}^{t}]_{j=1}^{|Y_i^t|}$ that summarises video $V_i^t$, where $y_{i,j}^{t}$ is the $j$-th token in the caption.

\subsection{Short-form Video Learning}
\label{sec:short-learning}
First we develop a primary short-form video summary model that identifies key information and summarises it. Inspired by multi-level annotations in the Ego4D-HCap dataset~\cite{islam2024video}, we divide Ego4D videos into 3-minute segments $V_i^s$ (which we use as the source domain $\mathcal{D}_s$). These segments have detailed descriptions $C_i^s$ for every 4 seconds of video from Ego4D, and captions $Y_i^s$ for the entire 3-minute segment from Ego4D-HCap.
During training, the model receives 3-minute video segments $V_i^s$ along with the corresponding 4-second descriptions $C_i^s$. The model learns to summarise 3-minute segments in a supervised manner using the segment captions $Y_i^s$.
We encode video segments $V_i^s$ using the frozen temporal visual transformer TimeSFormer~\cite{bertasius2021space} as feature encoder $F$. A visual-language alignment module $A_{vl}$ then aligns semantics between $V_i^s$ and $C_i^s$. Finally, a language model as text decoder $D$ generates 3-minute segment descriptions as predictions. This model is trained with 3-minute segment descriptions $Y_i^s$ (See Fig.\ref{fig:framework}(a)). We use cross-entropy and contrastive loss as follows:
\begin{small}
\begin{equation}
\begin{aligned}
\label{eq:short-term}
L_{\mathrm{pre}}&(V^s, C^s, Y^s) = - \sum_{j=1}^{J}{\log P(y_j^s=Y_{j}^s|y_{<j}^s,V^s, C^s)} \\&-\frac{1}{2N} \sum_{a=1}^{2N} \log \left(\frac{\exp\left(\frac{z_a^T z_{a^+}}{\tau}\right)}{\sum_{\substack{b=1 \\ b \neq a}}^{2N} \exp\left(\frac{z_a^T z_b}{\tau}\right)}\right),
\end{aligned}
\end{equation}
\end{small}
where $J$ is the number of tokens in $Y_j$; $z_a$ is a clip's feature vector and $z_a^+$ is the one from the immediate next clip, which is a positive sample; $z_b$  are feature vectors from other clips in the same batch, which we use as negative samples; $\tau$ is a scalar, and $N$ is the batch size. The first term in Eq.~\ref{eq:short-term} is the cross-entropy loss, which helps the model learn video summary. The second term is a temporal contrastive learning loss comparing sequential clips, helping the model find semantic connections between clips without supervision. 

\subsection{Hour-long Video Summary with LLMs}
The training in Sec.~\ref{sec:short-learning} gives the model basic video summary capabilities. However, directly applying this model to summarise hour-long videos yields poor results due to a domain gap caused by the difference in video length between the training source data and the target videos. We sidestep this difficulty by first segmenting long target videos $V^t_i$ into a collection of 3-minute segments $\mathbf{V}_i^t = \{v^t_{i,m}\}_{m=1}^{M_i}$, where $M_i$ is the number of segments for video $i$. $\mathbf{V}_i^t$ can eliminate the semantic gap caused by the video length, ensuring that the source pre-trained model can effective summarise these target segments (i.e., a short video summary model struggles to summarise an entire football match but can effectively summarise the players' short-term behaviours.). This way we obtain a pseudo caption $\hat{c}^t_{i,m}$ for each segment $\mathbf{V}_i^t$ from the pre-trained model:
\begin{equation}
\label{eq:split_form}
\hat{c}^t_{i,m} = D(A_{vl}(F(v^t_{i,m}))),
\end{equation}
With Eq.~\ref{eq:split_form}, we convert hour-long videos into a collection of short-form segment descriptions $\widehat{C}_i^t = \{\hat{c}^t_{i,m}\}_{m=1}^{M_i}$. 
Our working assumption is that in hour-long videos, only a few key segments are relevant to the underlying narrative and identifying these segments from all the redundant information is s particularly challenging task.
Fortunately, this is one of the tasks that LLMs excel at, due to their text reasoning and reflection capabilities \cite{sun2023survey}. Still, one remaining challenge is that summaries from LLMs $\widehat{Y}_{i}^t$ can be sensitive to the input prompt $Pr_{i}^t$. In addition, the optimal prompt is unknown, and may even differ between models, model versions or even across different videos.
Inspired by studies of LLMs as optimisers~\cite{Yang2023LargeLM}, we propose a $K$-iteration meta-prompting setup with multiple LLMs for iteratively refining the prompt $Pr_{i}^t$, which includes three components: a generator, an evaluator and an optimiser (see the right half of Fig.~\ref{fig:framework}).\\
\noindent\textbf{Generator LLM.}
Given an initial prompt ${{Pr}}_{i,k}^t$ from the $k$-th iteration, the generator LLM produces a summary capturing key information within $\widehat{C}_i^t$:
\begin{equation}
\label{eq:prompt_form}
{\widehat{Y}}_{i, k}^t = LLM_{\mathrm{gen}} (\widehat{C}_i^t,,{Pr}_{i, k}^t).
\end{equation}
We can then iteratively refine the prompt to improve the summary, progressively distilling essential content from extensive textual input.
This way we leverage the LLM's ability to efficiently extract key insights from from long-form texts often containing rich but sparse information.

\noindent\textbf{Evaluator LLM.}
The evaluator LLM evaluates the relevance and accuracy of the summary produced by the generator during the current iteration ${Sr}_{i,k}^t$, given the captions $\widehat{C}_i^t$,  as follows:
\begin{equation}
\label{eq:Score_form}
\mathrm{S}_{i, k}^t = LLM_{\mathrm{eval}} (\widehat{C}_i^t,\, \widehat{Y}_{i, k}^t).
\end{equation}
This assesses how well the summary $\widehat{Y}_{i,k}^t$ captures the essence of the captions, guiding the optimisation of the prompt in the next step. A higher score indicates that the summary better reflects the key information in the captions.\\
\noindent\textbf{Optimiser LLM.} 
The optimiser refines the generator's prompt ${Pr}_{i, k+1}^t$ using all previously evaluated prompts and their evaluation scores. We optimise the prompt by minimizing the domain gap and improving the summary in the next iteration:
\begin{equation}
\label{eq:Sn_form}
{Pr}_{i,k+1}^t = LLM_{\mathrm{opt}} (({Pr}_{i, 0}^t, \mathrm{S}_{i,0}^t), ..., ({Pr}_{i, k}^t, \mathrm{S}_{i,k}^t)).
\end{equation}

We evaluated several combinations of LLMs in meta-prompting (see in Tab. \ref{tab:meta-prompting}) and in our final implementation we use OpenAI's GPT-3.5-Turbo~\cite{openai2021gpt3.5} as the optimiser and generator, and Google's Gemini-1.5-Flash~\cite{reid2024gemini} as the evaluator. Using different LLMs helps reduce biases that might come from the LLMs and could be amplified during the optimisation process. Note that while some basic parts of the task description in $Pr_{i,k}^t$ stay fixed, the flexible elements are updated with each iteration, improving the summary $\widehat{Y}_{i,k}^t$. 
\\
\noindent\textbf{Early stopping strategy.} We iteratively refine the meta-prompting for each long video. However, given that videos differ in length and content, not all videos need the same number of iterations. If the summary score does not improve after a few iterations $l$ ($l<K$), we stop refining the prompt.

\subsection{Hour-long Video Adaptation}
After obtaining an optimised set of pseudo-summaries $\widehat{Y}^t$ for hour-long videos $V^t$, we use them to fine-tune the source pre-trained summary model. We also use the pseudo-descriptions $\widehat{C}^t$ as inputs for 3-minute video segments, so at this stage, we no longer rely on human annotations and source data.
To account for potential inaccuracies in the pseudo-summaries $\widehat{Y}^t$, we adopt a learning-with-noisy-labels strategy using a symmetric cross-entropy (SCE) loss~\cite{wang2019symmetric} as follows:
\begin{small}
\begin{equation}
\begin{aligned}
\label{eq:sce-term}
    &L_{\mathrm{SCE}}(V^t, \widehat{C}^t, \widehat{Y}^t) = L_{\mathrm{CE}}(V^t, \widehat{C}^t, \widehat{Y}^t) + L_{\mathrm{RCE}}(V^t, \widehat{C}^t, \widehat{Y}^t), \\
    & = -\sum_{j=1}^J \sum_{w=1}^W {P}(y_j^t = w \mid y_{<j}^t, V^t, \widehat{C}^t) \log \tilde{p}(w) \\
        & \quad - \sum_{j=1}^{J}{\log P(y_j^t=\widehat{Y}_{j}^t|y<j,\widehat{Y}^t)},
\end{aligned}
\end{equation}
\end{small}
where ${P}(y_j^t = w \mid y_{<j}^t, V^t, \widehat{C}^t) $ is the probability predicted by the model that the $j$-th token in the summary is $w$, given the previous output 
$y_{<j}^t $ and the inputs, and $\tilde{p}(w)$ is the token probability of $w$ in pseudo annotations $\widehat{Y}^t$, represented as a smoothed one-hot vector.

\subsection{Algorithm Overview}

Our framework follows a three-stage pipeline, which is summarised in Algorithm~\ref{alg:video_meta_prompting}. Rather than treating hour-long video summarisation as a monolithic task, we decompose it into modular stages that progressively bridge the gap between short-form supervised learning and long-form unsupervised video summarisation.
As shown in Fig.~\ref{fig:framework}, we first learn fine-grained grounding from labelled short-form videos, then use an LLM-based meta-prompting mechanism to generate pseudo-summaries for unlabelled long videos, and finally train a long-form summarisation model using these pseudo-summaries as supervision. This design enables unsupervised adaptation to hour-long videos without requiring costly annotations, while effectively leveraging both visual grounding and LLM's summarisation ability. 

\begin{algorithm}
\caption{Iterative Meta-Prompting for Hour-long Video Summarisation}
\label{alg:video_meta_prompting}
\begin{algorithmic}[1]
\STATE \textbf{Input:} Labelled source short-form videos $V^s$; Unlabelled target hour-long video $V^t$

\STATE \textbf{Step 1: Short-form Video Learning}

\STATE Extract visual features using frozen visual transformer $F$.
\STATE Train the summary model with video $V^S$ and its fine-grained captions $C_i^s$ (one every 4 seconds) as input, and the overall short-form summary $Y^s$ as supervision, using the alignment module $A_{vl}$ and language model $D$, optimised by loss $\mathcal{L}_{pre}$.
\STATE Obtain a short-form video summary model.

\STATE \textbf{Step 2: Pseudo Summary Generation}
\STATE Divide unlabelled target video $V^t$ into short segments $\{v_1^t, v_2^t, \dots, v_m^t\}$.
\STATE Generate segment pseudo-descriptions $\tilde{C}^t = \{C_1^t, C_2^t, \dots, C_m^t\}$ using frozen models ${F, A_{vl}, D}$.
\STATE Initialize prompt ${Pr}^t$ based on generated pseudo-descriptions $\tilde{C}^t$.
\FOR{iteration $k = 1, 2, \dots, K$}
\STATE \textbf{Generator}: Generate summary $\hat{Y}_k^t$ from prompt $Pr_k^t$.
\STATE \textbf{Evaluator}: Evaluate summary and output score $S_k^t$.
\STATE \textbf{Optimizer}: Optimize prompt to obtain improved prompt $Pr_{k+1}^t$.
\IF{Converged \textbf{or} scores $S_k^t$ and $S_{k-1}^t$ show no improvement over two consecutive iterations}
\STATE Break
\ENDIF
\ENDFOR
\STATE Obtain optimized summary $\hat{Y}^t$.

\STATE \textbf{Step 3: Hour-long Video Adaptation}
\STATE Fine-tune modules ${A_{vl}, D}$ using video $V^t$, optimized pseudo-summary $\hat{Y}^t$ with $L_{SCE}$.
\STATE \textbf{Output:} Model for summarising hour-long videos
\end{algorithmic}
\end{algorithm}

\section{Theoretical Analysis}
Suppose we have a collection of unlabelled hour-long videos \({\mathcal{D}_t}\), that are annotated with pseudo-captions \(\widehat{Y}^t\) via meta-prompting, as outlined in the previous section. Then \({\mathcal{D}_t}\) can be partitioned into two subsets: \(\mathcal{D}_{t}^+\), which includes videos with accurate pseudo-captions, and \(\mathcal{D}_{t}^-\), which contains videos with noisy pseudo captions (i.e., summaries that are misaligned with the video content, incomplete or incorrect). Let $\eta = \frac{|\mathcal{D}_t^-|}{|\mathcal{D}_t|}$ denote the label noise rate. We assume that samples from \(\mathcal{D}_{t}^+\) and \(\mathcal{D}_{t}^-\), denoted by \(\mathcal{U}_{t}^+\) and \(\mathcal{U}_{t}^-\) respectively, are independent and identically distributed (i.i.d.) of size \(m\) each.

\begin{table*}[ht]
    \centering
    \caption{Domain adaptation results to Ego4D-HCap dataset}
    \vspace{-10pt}
    \label{tab:results_ego4dHcap}
    \begin{adjustbox}{max width=\textwidth}
    \begin{tabular}{l|ccccccccccc}
        \midrule
        \multicolumn{12}{c}{\textbf{Ego4D-HCap \cite{islam2024video}  results}} \\
        \midrule
        {{\centering{\multirow{2}{*}{Model}}}} & {\multirow{2}{*}{Video Encoder}} & {\multirow{2}{*}{Text Decoder}} & {\multirow{2}{*}{Human Ann.}} 
         & {\multirow{2}{*}{Hier. Train}} & {\multirow{2}{*}{Train Params.}}& \multicolumn{3}{c}{Segment Description} & \multicolumn{3}{c}{Video Summary} \\
        \cmidrule(lr){7-9} \cmidrule(lr){10-12}
        & & & & & & C & R & M & C & R & M \\
        \midrule
        \multicolumn{12}{c}{Fully supervised} \\\hline
        LaViLa  \cite{zhao2023learning} & TSF-B & GPT2 & \checkmark & \texttimes & 258M & 24.6 & 33.3 & 15.3 & 6.5 & 24.0 & 11.0 \\
        LaViLa \cite{zhao2023learning}+GPT2 \cite{lagler2013gpt2} & TSF-B & GPT2 & \checkmark  & \texttimes & 336M & 38.2 & 38.1 & 16.6 & 18.0 & 29.5 & 12.8 \\
        LaViLa \cite{zhao2023learning}+FLANT5 \cite{chung2024scaling} &TSF-B & FT5-XL &  \checkmark & \texttimes & 586M & 39.1 & 38.8 & 16.9 & 20.1 & 30.1 & 13.2 \\
        Video ReCap \cite{islam2024video} &TSF-B & GPT2 & \checkmark & \checkmark & 113M & \textbf{46.9} & \textbf{39.7} & \textbf{18.6} & \textbf{29.3} & \textbf{32.6} & \textbf{14.2} \\\hline
        \multicolumn{12}{c}{Unsupervised} \\
        \midrule
        \textbf{Zero-Shot} \\
        BLIP2 \cite{li2023blip}+GPT3.5 \cite{openai2021gpt3.5} &VIT-G & FT5-XL &  \texttimes & \texttimes & 0 & 5.7 & 16.9 & 13.5 & 11.1 & 22.4 & 12.1 \\
        LaViLa \cite{zhao2023learning}+GPT3.5 \cite{openai2021gpt3.5}  &TSF-B & GPT2 & \texttimes & \texttimes & 0 & 5.8 & 19.8 & 13.5 & 12.2 & 24.5 & 12.5 \\
        \midrule
        \textbf{Domain Adaptation} \\
        VisMaP (ours) & TSF-B & GPT2 & \texttimes & \texttimes & 113M & \textbf{45.6} & \textbf{39.6} & \textbf{18.6} & \textbf{26.0} & \textbf{29.9} & \textbf{13.1}\\\hline
    \end{tabular}
    \end{adjustbox}
\end{table*}

\noindent\textbf{Theorem 1.}
\textit{Let \(h\) be a captioning hypothesis trained on \({\mathcal{D}_t}\). With at least \(1 - \delta\) probability, the true error \(\epsilon_t(h)\) is bounded by:}
\begin{small}
\begin{equation}
\begin{aligned}
\epsilon_t(h) &\leq 2\epsilon_{t}^+(h) + \frac{1}{2} d_{H\Delta H}(\mathcal{D}_{t}^+, \mathcal{D}_{t}^-) \\&+ 2\left( {\frac{d \log(2m) + \log\left(\frac{2}{\delta}\right)}{m}} \right)^{\frac{1}{2}} + \lambda,
\end{aligned}
\label{eq:error-bound}
\end{equation}
\end{small}
\textit{where {\small{$$\lambda = \mathbb{E}_{\mathcal{D}_{t}^+}\left[\sum_{k \neq y} \eta_{yk} \ell_{\mathrm{rce}}(f(v), k) 
- \eta_y \ell_{\mathrm{rce}}(f(v), y)\right]$$}} is the expected additional loss due to label noise.}
\noindent The left-hand side term in Eq.~\ref{eq:error-bound}, $2\epsilon_t^+(h)$, is constrained by supervised training on correctly labelled pseudo-captions. The second and third terms capture the domain divergence, and are regulated through the second stage of the long video summary using LLMs. The last term $\lambda$, is governed by the SCE loss in Eq.~\ref{eq:sce-term} during the target adaptation phase, ensuring effective learning under noisy label conditions. 

\noindent\textbf{Proof.} Consider the unlabelled video domain \(\mathcal{D}_t\), which is partitioned into two subsets: \(\mathcal{D}_t^+\), containing accurately pseudo-captioned videos, and \(\mathcal{D}_t^-\), containing noisy pseudo-captioned videos. The total error on the target domain can be expressed as:
\begin{equation}
\epsilon_t(h) = \mathbb{E}_{v \in \mathcal{D}_t}[|h(v)-f(v)|] = \epsilon_t^+(h) + \epsilon_t^-(h).
\end{equation}

By adding and subtracting appropriate terms and applying the triangle inequality, we derive:
\begin{equation}
\begin{aligned}
\epsilon_t(h) &= \epsilon_t^+(h) + \epsilon_t^-(h) + \epsilon_t^+(h) - \epsilon_t^+(h) \\
&\quad + \epsilon_t^+(h, f^-) - \epsilon_t^+(h, f^-) \\
&\leq 2\epsilon_t^+(h) + |\epsilon_t^-(h, f^-) - \epsilon_t^+(h, f^-)| \\
&\quad + |\epsilon_t^+(h, f^-) - \epsilon_t^+(h, f^+)|,
\end{aligned}
\end{equation}
where \(f^+\) and \(f^-\) denote the true and noisy target functions, respectively.

Using the \(\mathcal{H} \Delta \mathcal{H}\)-divergence \cite{ben2010theory}, we have:
\begin{equation}
|\epsilon_t^-(h, f^-) - \epsilon_t^+(h, f^-)| \leq \frac{1}{2} d_{\mathcal{H} \Delta \mathcal{H}}(\mathcal{D}_t^+, \mathcal{D}_t^-).
\end{equation}

Given empirical samples \(\mathcal{U}_t^+\) and \(\mathcal{U}_t^-\), and applying VC-dimension generalisation bounds, with probability at least \(1 - \delta\), we obtain:
\begin{equation}
\begin{aligned}
\epsilon_t(h) &\leq 2\epsilon_t^+(h) + \frac{1}{2} d_{\mathcal{H} \Delta \mathcal{H}}(\mathcal{U}_t^+, \mathcal{U}_t^-) \\
&\quad + 2\left({\frac{d \log (2m) + \log \left(\frac{2}{\delta}\right)}{m}}\right)^{\frac{1}{2}} \\
&\quad + |\epsilon_t^+(h, f^-) - \epsilon_t^+(h, f^+)|,
\end{aligned}
\end{equation}
where \(d\) is the VC-dimension of the hypothesis class \(\mathcal{H}\).

The final term quantifies the expected loss introduced by asymmetric label noise. Evaluated using the symmetric cross-entropy (SCE) loss, it corresponds to:
\begin{equation}
|\epsilon_t^+(h, f^-) - \epsilon_t^+(h, f^+)| 
= \mathbb{E}_{(v, y) \sim \mathcal{D}_t^+} \left[ \left| \ell(f(v), \tilde{y}) - \ell(f(v), y) \right| \right].
\end{equation}
Thus, we recover the bound in Theorem 1, completing the proof. A more detailed derivation is in Supplementary Material.

\section{Experiments}
\subsection{Evaluation, Data and Testing Assumptions}
\noindent We evaluate our approach in three scenarios:
(1) We evaluate VisMaP’s summarisation ability under a video length gap using the Ego4D-HCap~\cite{islam2024video} dataset. (2) We measure VisMaP's ability to bridge cross-domain gaps under varying content conditions on the short-form video caption datasets MSRVTT~\cite{xu2016msr}, MSVD~\cite{khan2020hybrid} and YouCook2~\cite{zhou2018towards}. (3) We test VisMap's (trained on hour-long videos) cross-length generalisation ability on the short-form video dataset EgoSchema~\cite{mangalam2023egoschema}, which contains videos that are on average 3 minutes long.  

\noindent\textbf{Ego4D-HCap}~\cite{islam2024video} is an hour-long video dataset with three levels of detail: second-length clips, minute-length segments, and full videos. We use the annotated minute-length video segments as the source domain $\mathcal{D}_s$ and the unlabelled full videos as the target domain $\mathcal{D}_t$ 
We compare VisMaP with Video ReCaP \cite{islam2024video}, the only model specifically designed for hour-long video summaries, and LaViLa~\cite{zhao2023learning}, a video captioning model that can be extended for long-form videos, both of which were trained on the Ego4D-HCap dataset. The Video Recap model uses additional 4-second video clips for training.
We also compare VisMaP against zero-shot approaches using BLIP2~\cite{li2023blip} combined with GPT 3.5~\cite{openai2021gpt3.5} and
  LaViLa with GPT 3.5. We generated captions with BLIP2 and LaViLa respectively, and then processed these captions with GPT 3.5 to generate the summaries.  
We also compared VisMaP with LaViLa+GPT2~\cite{lagler2013gpt2}, LaViLa+FLAN-T5~\cite{chung2024scaling}, both are fine-tuned on annotated 3-minute video segments and full videos.

\noindent\textbf{MSRVTT}~\cite{xu2016msr}, \textbf{MSVD}~\cite{khan2020hybrid} and \textbf{Youcook2}~\cite{zhou2018towards} are video captioning datasets featuring clips from various scenes, usually under 5 minutes long, all shot from a third-person perspective.  We use these videos to evaluate the semantic generalisation ability of VisMaP (source pre-trained on 4-second video clips from the Ego4D-HCap dataset) under similar duration settings. We also compared VisMaP with leading supervised methods~\cite{luo2020univl, wang2022git,lin2022swinbert,zhang2023simple, he2024ma}.

\noindent\textbf{EgoSchema}~\cite{mangalam2023egoschema} includes over 5,000 curated multiple-choice questions and answers over real-world videos with average duration of 3 minutes. We use this dataset to evaluate how well VisMaP (trained on hour-long videos) can perform on shorter video clips. We segment videos into various clip lengths to generate hierarchical detailed video captions at different levels. We feed these captions into GPT3.5/4 to produce the final summary. We compared VisMaP with leading methods~\cite{lin2022egocentric,pramanick2023egovlpv2,wang2022internvideo, ye2023mplug} to highlight its adaptability and performance across various video formats and topical complexities.

\begin{table*}[t]
\centering
\begin{minipage}{0.45\textwidth}
\centering
\caption{ Short-form VideoQA on EgoSchema}
\label{tab:ego}
\vspace{-10pt}
\renewcommand{\arraystretch}{1.3}
\begin{adjustbox}{max width=\textwidth}
\begin{tabular}{@{}lcccc@{}}
\toprule
{\multirow{2}{*}{Model}}                 & Input  & Ego4D & QA  \\ 
& Feature & Pretain & Acc \\
\midrule
GPT3.5 \cite{openai2021gpt3.5}           & Question       & \texttimes     & 19.6  \\\hline
FrozenBiLM \cite{yang2022zero} & Video          & \texttimes     & 26.9   \\
mPLUG-Owl \cite{ye2023mplug}         & Video          & \texttimes     & 31.1   \\
InternVideo \cite{wang2022internvideo}       & Video          & \texttimes     & 32.1   \\\hline
EgoVLP \cite{lin2022egocentric}           & Video          & \checkmark     & 34.9  \\
EgoVLPv2 \cite{pramanick2023egovlpv2}         & Video          & \checkmark     & 34.2  \\\hline
LaViLa \cite{zhao2023learning}+GPT3.5 \cite{openai2021gpt3.5}  & Captions       & \checkmark     & 44.3  \\
Video ReCap \cite{islam2024video} + GPT3.5 \cite{openai2021gpt3.5} & Hier. Captions & \checkmark & 50.2  \\\hline
VisMaP + GPT3.5 \cite{openai2021gpt3.5} & Hier. Captions & \checkmark & 52.5\\
VisMaP + GPT4 \cite{achiam2023gpt} & Hier. Captions & \checkmark & \textbf{53.4}\\
\bottomrule
\end{tabular}
\end{adjustbox}
\end{minipage}%
\hfill
\begin{minipage}{0.55\textwidth}
\centering
\caption{Short-form Video Captioning Results}
\label{tab:results_short_captioning}
\vspace{-10pt}
\begin{adjustbox}{max width=\textwidth}
\begin{tabular}{lcccccc}
\toprule
\multicolumn{7}{c}{\textbf{Short-form Video Captioning}} \\
\midrule
\multirow{2}{*}{Model} & \multicolumn{2}{c}{MSR-VTT} & \multicolumn{2}{c}{MSVD} & \multicolumn{2}{c}{YouCook2} \\
\cmidrule(lr){2-3} \cmidrule(lr){4-5} \cmidrule(lr){6-7}
& M & C & M & C & M & C \\
\midrule
\textbf{Supervised} \\
UniVL \cite{luo2020univl} & 28.2 & 49.9 & 29.3 & 52.8 & – & 127.0 \\
SwinBERT \cite{lin2022swinbert} & 29.9 & 53.8 & 41.3 & 120.6 & 15.6 & 109.0 \\
GIT \cite{wang2022git} & 32.9 & 73.9 & 51.1 & 180.2 & 17.3 & 129.8 \\
Video-LLaMA \cite{zhang2023video} & 32.9 & 71.6 & 49.8 & 175.3 & 16.5 & 123.7 \\
MA-LMM \cite{he2024ma}  & \textbf{33.4} & \textbf{74.6} & \textbf{51.0} & \textbf{179.1} & \textbf{17.6} & \textbf{131.2} \\
\midrule
\textbf{Unsupervised} \\
Video ReCap \cite{islam2024video} & 7.1 & 5.5 & 10.1 & 8.2 & 7.5 & 8.9 \\
VisMaP (Ours) & \textbf{29.1} & \textbf{52.8} & \textbf{38.9} & \textbf{113.3} & \textbf{15.9} & \textbf{117.1} \\
\bottomrule
\end{tabular}
\end{adjustbox}
\end{minipage}
\vspace{-4pt}
\end{table*}

\begin{table}[t]
    \caption{Ablation study of different modules. Here TPL means ``target pseudo labelling'', SCL is ``source contrastive learning'', CSG is ``cycle summary generation'', CPG is ``cycle prompt generation'', and SCE is ``Symmetric Cross Entropy''.}
    \vspace{-10pt}
    \label{tab:ablation}
    \centering
    \begin{adjustbox}{max width=0.46\textwidth}
    \begin{tabular}{c|ccccc|ccc}\hline
    \multicolumn{6}{c|}{Model's variants} & \multicolumn{3}{c}{Ego4D-HCap}\\\hline
          & TPL & SCL & CSG & CPG & SCE & C & R & M \\\hline
          1 & \checkmark & & & & & 22.6 & 26.9 & 11.7 \\
          2 & \checkmark & \checkmark & & & & 24.5 & 28.0 & 11.8 \\
          3 & \checkmark & \checkmark & \checkmark & & & 24.6 & 28.8 & 12.5 \\
          4 & \checkmark & \checkmark & \checkmark & \checkmark & & 25.8 & 29.8 & 12.8 \\
          5 & \checkmark & \checkmark & \checkmark & \checkmark & \checkmark & \textbf{26.0} & \textbf{29.9} & \textbf{13.1} \\\hline     
        \end{tabular}
        \end{adjustbox}
        \vspace{-10pt}
\end{table}

\noindent\textbf{Metrics.}
To measure performance across all tests, we rely on three widely-used captioning metrics: CIDEr (C)~\cite{vedantam2015cider}, which quantifies the similarity between a candidate's captions and reference captions based on the consensus of $n$-grams; ROUGE-L (R)~\cite{lin2004rouge}, which measures the longest common subsequence between the generated text and the reference to assess fluency and text coherence; and METEOR (M)~\cite{banerjee2005meteor}, which evaluates translation hypotheses by aligning them to reference translations based on their semantic and syntactic features. For the EgoSchema dataset, we use QA-accuracy to evaluate on the VQA task. Higher scores on these metrics indicate better performance.

\noindent\textbf{Implementation Details.} Our model is trained on one NVIDIA A100 GPU for 25 epochs during each of the two phases. A frozen TimeSformer~\cite{bertasius2021space} is used as an encoder to generate video embeddings. DistilBERT~\cite{sanh2019distilbert}, a language model, aligns video embeddings with text inputs. We aggregate the 4-second embeddings from TimeSFormer to form a fixed-length embedding for various video lengths. A GPT2 text decoder is used for the final output. We used the Adam optimiser~\cite{diederik2014adam} with a learning rate of 3e-5 and a weight decay of 0.01. We used GPT 3.5 Turbo as the optimiser and generator, Gemini 1.5 Flash as the evaluator, with 5 cycles meta-prompting to generate pseudo-summary (see Fig.~\ref{fig:framework}).

\begin{table}[t]
    \centering
    \caption{Effect of iteration count on pseudo label generation.}
    \vspace{-10pt}
    \label{tab:num_iter}
    \centering
    \begin{adjustbox}{max width=1.0\textwidth}
    \begin{tabular}{c|ccc}\hline
         iteration $K$ & C & R & M \\\hline
         1 & 24.5 & 28.0 & 11.8 \\
         2 & 24.9 & 28.6 & 12.2 \\
         3 & 25.1 & 29.4 & 12.6 \\
         4 & 25.8 & 29.6 & 12.9 \\
         5 & \textbf{26.0} & 29.9 & \textbf{13.1} \\
         6 & 25.8 & \textbf{29.9} & 13.0 \\\hline
    \end{tabular}
    \label{tab:my_label}
    \end{adjustbox}
\end{table}

\begin{table}[t]
    \caption{Evaluating LLMs in meta-prompting.}
    \centering
    \vspace{-10pt}
    \label{tab:meta-prompting}
    \begin{adjustbox}{max width=0.5\textwidth}
    \begin{tabular}{cccc|ccc}\hline
    \multicolumn{4}{c|}{Four LLM combinations  } & \multicolumn{3}{c}{Ego-4D HCap}\\\hline
           & Optimiser  & Evaluator & Generator  & C & R & M \\\hline
          1 & Gemini \cite{reid2024gemini} & GPT3.5 \cite{openai2021gpt3.5} & 
           Mixtral \cite{jiang2024mixtral} & 24.9 & 27.9 & 11.6 \\
           2 &  GPT3.5 & 
            Gemini  & 
           Mixtral & 24.7 & 27.6 & 11.4 \\
           3 &  Mixtral  & GPT3.5  & Mixtral  & 24.6 & 27.9 & 11.5 \\
           4 &  GPT3.5 & GPT4 \cite{achiam2023gpt} & GPT3.5 & 25.4 & 29.3 & 12.5 \\
           5 &  GPT3.5  & GPT3.5  & GPT3.5  & 25.6 & 29.5 & 12.6 \\
           6 &  GPT4 & GPT4  & GPT4 & 25.7 & 29.4 & 12.6 \\
           7 & GPT3.5 & Gemini & GPT3.5 &\textbf{26.0} & \textbf{29.9} & \textbf{13.1} \\\hline  
        \end{tabular}
    \end{adjustbox}
\end{table}

\subsection{Results}
\label{sec:results}

\noindent\textbf{Results on Ego4D-HCap.}
Tab.~\ref{tab:results_ego4dHcap} shows the performance of VisMaP compared to supervised and unsupervised methods on the Ego4D-HCap dataset.
Overall, zero-shot methods perform significantly worse than supervised methods, which underscores the complexity of long-form summarisation. 
Our results show that VisMap outperforms methods that use video segments and full videos for supervised learning. The one exception is the state-of-the-art Video Recap, which outperforms VisMap in summarisation because it is fully supervised; even then, VisMap's performance is comparable, which is impressive given that it is an unsupervised method. These results show that using LLM-generated pseudo-summaries with meta-prompting is an effective strategy for long-form video summarisation, especially given that it sidesteps the need for costly annotations.

\begin{figure*}[ht]
   \centering  
   \vspace{10pt}
   \includegraphics[width=18cm, trim=1.1cm 0 0 0, clip]{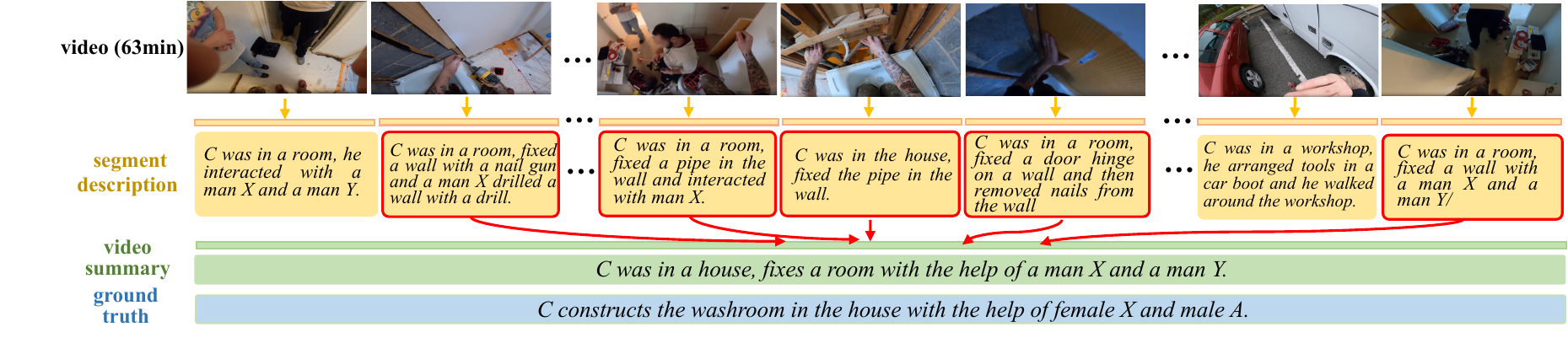} 
   \vspace{-20pt}
   \caption{An example of summaries from ViSMaP on the Ego4D-HCap dataset. 
   }
   \label{fig:vis}
   \vspace{5pt}
\end{figure*}

\noindent\textbf{Result on EgoSchema.} Tab.~\ref{tab:ego} shows the evaluation of VisMap trained on hour-long videos directly on the EgoSchema dataset (average duration of 3-minutes) to evaluate how well our model can generalise going from hour-long videos to short videos. VisMaP significantly outperforms long-form video methods and classical approaches, further proving its superior performance and generalisation ability across various video analysis tasks. The consistent strong performance across datasets highlights our method’s robustness and adaptability to diverse videos and tasks.

\noindent\textbf{Results on short video captioning.} Tab.~\ref{tab:results_short_captioning} shows short video captioning evaluation on the MSRVTT, MSVD, and YouCook2 datasets. We pit unsupervised VisMaP against supervised learning approaches. Results show that VisMaP can effectively caption short videos with significant differences in perspective and content, and achieves results that are close to or even surpass supervised models.

\noindent\textbf{Qualitative Results on the Ego4D-HCap dataset.}
Fig.~\ref{fig:vis} shows an example of the type of summaries that VisMap can generate without supervision on a 63 minute long video from the Ego4D-HCap dataset. Following the flow in Fig.~\ref{fig:framework}, the source summary model, which was trained on short-form videos, produces captions for each 30-second long segment. The collection of segment descriptions for this video has a total of 3480 words. After adaptation, the target model, which is adjusted to content variations, selects the key information (shown in red boxes in the figure) to create an accurate and concise summary that is specific to the input video (third row in Fig.~\ref{fig:vis}), which in this example is 19 words long. These summaries can be readily compared to ``ground truth'' summaries generated by human annotators (bottom row in Fig.~\ref{fig:vis}).
\noindent\textbf{Module Analysis.} To understand the contribution of the constituent components of our method to its performance we carry out an ablation study. The first row
of Tab.~\ref{tab:ablation} contains a basic baseline approach that uses a model pretrained on the source domain, and employs GPT-3.5 to create summaries. We use these summaries as pseudo-labels to train a new captioning model. This baseline, similar to LLoVi~\cite{zhang2023simple}, is less effective because it cannot accurately identify key information, resulting in summaries filled with unnecessary details. 
Row 2 contains a model enhanced by incorporating source contrastive learning during pre-training, which enables the extraction of information from the source domain, and improves the accuracy of the target pseudo-summaries.
Rows 3 and 4 contain models that introduce cycle summary and prompt generation strategies, respectively. These techniques use meta-learning to iteratively improve summaries and prompts, leading to more accurate pseudo-summaries and validating our meta-prompting approach. Finally, row 5 adds the SCE loss during target adaptation to reduce the impact of noisy labels and ensure effective semantic learning.

\noindent\textbf{Analysis of the number of iterations.} In Tab.~\ref{tab:num_iter}, we present an evaluation of the impact that the number of iterations $K$ has on performance. These results clearly show the benefit from each iteration up to the fifth, after which it stabilises. Therefore, we have set $K=5$ in our evaluations.

\noindent\textbf{LLMs in meta prompting.} To understand which combination of LLMs as optimiser, evaluator and generator works best, we evaluate different combinations and report our results in Tab.~\ref{tab:meta-prompting}. The results reveal that different LLM choices significantly impact performance. While GPT-4 demonstrates strong overall capabilities, it does not consistently outperform all configurations. Notably, the combination of GPT-3.5 as both the optimiser and generator, with Gemini as the evaluator, achieves the highest scores across all metrics, suggesting that Gemini provides effective evaluation feedback while GPT-3.5 generates outputs that align well with the evaluation criteria. In contrast, Mixtral-based setups generally yield lower performance, indicating that its effectiveness may be role-dependent. These findings highlight the importance of selecting LLMs based on their specific strengths rather than defaulting to a single powerful model for all tasks.

\noindent\textbf{Supplementary material}
 Sections S1 and S2 provide code information and implementation details. Section S3 presents an in-depth theoretical analysis, Sections S4 and S5 contain an analysis of the LLMs in meta-prompting and detailed limitations.

\section{Conclusion}
We introduce ViSMaP, an unsupervised method for summarising hour-long videos by leveraging annotated short-video datasets and a meta-prompting strategy. We first generate high-quality summaries via meta-prompting and then train an end-to-end summarisation model, reducing reliance on extensive annotations. Experiments show ViSMaP achieves performance comparable to fully supervised methods and adapts effectively to diverse video datasets.
One limitation of ViSMaP is that it relies on pseudo labels generated by a source-domain model, which may limit performance under large domain shifts. The current approach also uses only visual information, without incorporating audio or transcripts, which can affect summary quality. TFuture work can explore multimodal inputs, hierarchical summarisation at different resolutions, and more generalisable meta-prompting methods for broader use.

\bibliographystyle{ACM-Reference-Format}
\bibliography{bibliography}

\newpage

\title{Supplementary Materials for ViSMaP: Unsupervised Hour-long Video Summary by Meta-Prompting}


\maketitle
\tableofcontents 
\section{Code Release}
The code will be released after our paper is accepted.

\section{More Implement Details}
\noindent\textbf{Module details.} Our experiments (training and evaluation) are conducted using Pytorch on a single NVIDIA A100 GPU. Specifically, the feature encoder is a TimeSformer, which processes videos resized to 224x224 pixels. This encoder inherits parameters from a model pre-trained as detailed in \cite{zhao2023learning}. During training, the visual encoder remains frozen, and we use the output from the final \texttt{cls} layer as the input for our visual language alignment module. In our Video-Language (VL) Alignment module, we employ DistilBERT, a pretrained 6-layer transformer encoder model. We maintain the self-attention blocks as frozen and introduce a trainable cross-attention module to each layer. The model processes video features from the video encoder and captions produced in the preceding hierarchy as inputs. We employ a pretrained GPT2 model as a text decoder, featuring a 12-layer transformer structure. Within each transformer layer, we incorporate a gated cross-attention block, training only these blocks while freezing the remaining components. Each block comprises a cross-attention layer and a feed-forward layer, enhanced by $\tanh$ gating initially set to zero.

\noindent\textbf{Training details.} The training of VisMaP involves three phases. The first phase trains on supervised video segments from the Ego4d-HCap dataset, with a 3-minute duration per segment. During segments-level training, the model received two inputs, fine-grained descriptions $C^s$ for every 4 seconds of video, along with a visual embedding for every 3 minute of video generated by the feature encoder after subsampling to a fixed dimension. The model is supervised with a textual description $Y^s$ for the entire 3-minute long video segment. The fine-grained video captions and visual embedding are then inputted into a VL Alignment module for cross-modal semantic alignment. The aligned embeddings are subsequently fed into a decoder, which acquires basic video summary and understanding capabilities from the ground-truth segment descriptions $Y^s$. This phase requires training for up to 50 epochs with a batch size of 64. In the second phase, all model parameters are frozen, and the model processes unlabeled target dataset videos, split into 3-minute segments. More specifically the pretrained model from the first phase is used to generate a collection of segment descriptions, one for each 3-minute segment of the long-form videos. These captions are then inputted into our meta-prompting with GPT-3.5 as both generator and optimizer, and Gemini-1.5 Flash as the evaluator, iterating through the prompt and summary optimization for up to a maximum of 5 iterations to produce pseudo-summaries for each long video in the target domain. We also employ early-stopping for our iterative prompt adaptation, whereby we terminate the loop if there is no improvement. The third phase involves using the pseudo-summaries generated by our iterative prompt adaptation process to train a full video summary model on unlabelled long-form videos. During this phase, the model inputs are segment descriptions for every 3 minutes of video, generated with the phase one segment-level model, and full video visual embeddings generated using the features encoder. The model inputs are fed into a VL alignment module and the resulting embeddings are then inputted into a decoder and supervised using the pseudo-summaries for up to 200 epochs, with the batch size set to 32. We use \texttt{AdamW} optimizer with a learning rate of $3e-5$ and weight decay of $0.01$. 

\begin{figure*}[tb]
   \centering  
   \includegraphics[width=18.55cm, trim=0.8cm 0 0 0, clip]{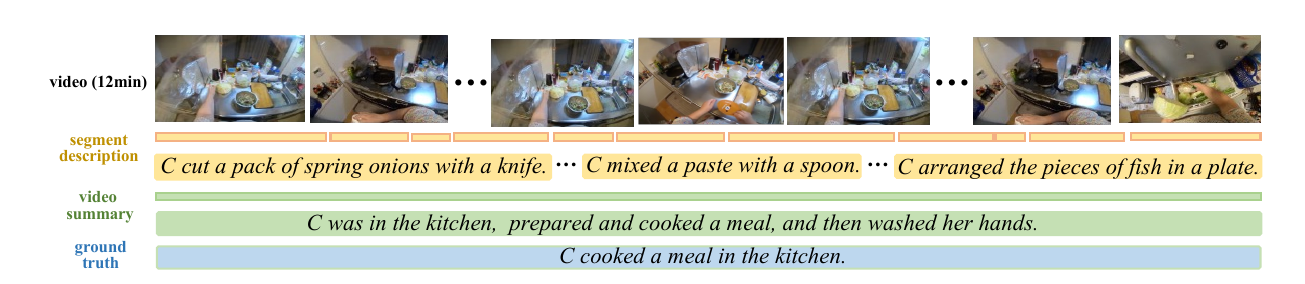} 
   \vspace{-20pt}
   \caption{Qualitative Results on Ego4D dataset.}
   \label{fig:vis1}
\end{figure*}

\section{Theoretical Details}
Hypothesize a function represented as $h : \mathcal{V} \rightarrow \{0,1\}$. The probability according to the distribution $\mathcal{D}$ that a hypothesis $h$ disagrees with a labeling function $f$ is defined as:
\begin{equation}
\epsilon(h,f) = \mathbb{E}_{v \sim \mathcal{D}}[|h(v) - f(v)|],
\end{equation}

\noindent\textbf{Theorem 1.} \textit{Consider an unlabeled hour-long video domain \( \mathcal{D}_t \), to which pseudo-captions \( \widehat{Y}^t \) are assigned by an initial captioning process. Then \( \mathcal{D}_t \) can be partitioned into two disjoint subsets:
\( \mathcal{D}_{t}^+ \): instances correctly labeled by the pseudo-captioning process,  \( \mathcal{D}_{t}^- \): instances labeled with noise by the pseudo-captioning process.
The label noise rate is \( \eta \).
Suppose we independently and identically sample (i.i.d.) \( m \) instances from \( \mathcal{D}_{t}^+ \) and \( \mathcal{D}_{t}^- \), forming samples \( \mathcal{U}_{t}^+ \) and \( \mathcal{U}_{t}^- \), respectively.
Let \( \ell(\cdot, \cdot) \) be a loss function applicable to a hypothesis and a dataset (for empirical error estimation) or a distribution (for generalization error estimation). Given a hypothesis \( h \) parameterized by \( \theta \), trained on \( \mathcal{D}_t \), and belonging to a hypothesis space \( \mathcal{H} \) with VC-dimension \( d \), then with probability at least \( 1 - \delta \), the following inequality holds:}
\begin{equation}
\begin{aligned}
\epsilon_t (h) &\leq 2\epsilon_{t}^+ (h) + \frac{1}{2} d_{\mathcal{H}\Delta \mathcal{H}}(\mathcal{D}_{t}^+, \mathcal{D}_{t}^-) \\
&+ 2\sqrt{\frac{d \log(2m) + \log\left(\frac{2}{\delta}\right)}{m}} + \lambda,
\end{aligned} 
\end{equation}
\textit{where \(\lambda = \mathbb{E}_{\mathcal{D}_{t}^+}\left[\sum_{k \neq y} \eta_{yk} \ell_{\mathrm{rce}}(f(v), k) - \eta_y \ell_{\mathrm{rce}}(f(v), y)\right]\) represents the expected additional loss due to label noise, and \( d_{\mathcal{H}\Delta \mathcal{H}}(\mathcal{D}_{t}^+, \mathcal{D}_{t}^-) \) denotes the distribution divergence.}

\noindent\textbf{Proof.} Recall that \( \epsilon_t(h) = \mathbb{E}_{v \in \mathcal{D}_t} [|h(v) - f(v)|] \), and \( \mathcal{D}_t = \{\mathcal{D}_{t}^+, \mathcal{D}_{t}^-\} \). Therefore:
\begin{equation}
\begin{aligned}
\epsilon_t(h) &= \epsilon_{t}^+(h) + \epsilon_{t}^-(h) + \epsilon_{t}^+(h) - \epsilon_{t}^+(h) \\
&\quad + \epsilon_{t}^+(h, f^-) - \epsilon_{t}^+(h, f^-) \\
&\leq 2\epsilon_{t}^+(h) + |\epsilon_{t}^-(h, f^-) - \epsilon_{t}^+(h, f^-)| \\
&\quad + |\epsilon_{t}^+(h, f^-) - \epsilon_{t}^+(h, f^+)|,
\end{aligned}
\end{equation}
where \( f^+ \) and \( f^- \) are the true and noisy labeling functions, respectively.

Following \cite{ben2010theory}, the \( \mathcal{H} \)-divergence between distributions \( D \) and \( D' \) over domain \( \mathcal{V} \) is:
\begin{equation}
d_\mathcal{H}(D, D') = 2 \sup_{h \in \mathcal{H}} | \Pr_{D}(I(h)) - \Pr_{D'}(I(h)) |,
\end{equation}
For any two hypotheses \( h, h' \in \mathcal{H} \):
\begin{equation}
\begin{aligned}
|\epsilon_s(h, h') - \epsilon_t(h, h')| &\leq \sup_{h, h' \in \mathcal{H}} |\epsilon_s(h, h') - \epsilon_t(h, h')| \\
&= \frac{1}{2} d_{\mathcal{H} \Delta \mathcal{H}}(D_s, D_t),
\end{aligned}
\end{equation}
Applying this, we have:
\begin{equation}
\begin{aligned}
\epsilon_t(h) &\leq 2\epsilon_{t}^+(h) + \frac{1}{2} d_{\mathcal{H} \Delta \mathcal{H}}(\mathcal{D}_{t}^+, \mathcal{D}_{t}^-) \\
&\quad + |\epsilon_{t}^+(h, f^-) - \epsilon_{t}^+(h, f^+)|,
\end{aligned}
\end{equation}
Letting \( \mathcal{U}_t^+, \mathcal{U}_t^- \) be samples from \( \mathcal{D}_{t}^+, \mathcal{D}_{t}^- \), and using VC generalisation bounds:
\begin{equation}
\begin{aligned}
\epsilon_t(h) &\leq 2\epsilon_{t}^+(h) + \frac{1}{2} d_{\mathcal{H} \Delta \mathcal{H}}(\mathcal{U}_t^+, \mathcal{U}_t^-) \\
&\quad + 2\sqrt{\frac{d \log (2m) + \log (\frac{2}{\delta})}{m}} \\
&\quad + |\epsilon_{t}^+(h, f^-) - \epsilon_{t}^+(h, f^+)|,
\end{aligned}
\end{equation}
Evaluating the final term using the noise model:
\begin{equation}
\begin{aligned}
|\epsilon_{t}^+(h, f^-) - \epsilon_{t}^+(h, f^+)| = \mathbb{E}_{\mathcal{D}_t^+} \left[ \sum_{k \neq y} \eta_{yk} \ell_{\mathrm{rce}}(f(v), k) - \eta_y \ell_{\mathrm{rce}}(f(v), y) \right] = \lambda
\end{aligned}
\end{equation}
Hence, the final bound becomes:
\begin{equation}
\epsilon_t(h) \leq 2\epsilon_{t}^+(h) + \frac{1}{2} d_{\mathcal{H}\Delta \mathcal{H}}(\mathcal{U}_t^+, \mathcal{U}_t^-) + 2\sqrt{\frac{d \log (2m) + \log (\frac{2}{\delta})}{m}} + \lambda
\end{equation}

\begin{table}[t]
    \caption{Evaluating LLMs in meta-prompting.}
    \centering
    \vspace{-10pt}
    \label{tab:meta-prompting}
    \begin{adjustbox}{max width=0.5\textwidth}
    \begin{tabular}{ccc|ccc}\hline
    \multicolumn{3}{c|}{Four LLM combinations
    } & \multicolumn{3}{c}{Ego-4D HCap}\\\hline
           Optimizer  & Evaluator & Generator  & C & R & M \\\hline
           GPT3.5 \cite{openai2021gpt3.5} & GPT4 \cite{achiam2023gpt} & GPT3.5 \cite{openai2021gpt3.5} & 25.39 & 29.30 & 12.47 \\
           Gemini \cite{reid2024gemini} & GPT3.5 \cite{openai2021gpt3.5} & Mixtral \cite{jiang2024mixtral} & 24.85 & 27.93 & 11.46 \\
           GPT3.5 \cite{openai2021gpt3.5}& Gemini \cite{reid2024gemini}  & Mixtral \cite{jiang2024mixtral} & 24.73 & 27.56 & 11.39 \\
           Mixtral \cite{jiang2024mixtral} & GPT3.5 \cite{openai2021gpt3.5} & Mixtral \cite{jiang2024mixtral} & 24.61 & 27.88 & 11.47 \\
           Mixtral \cite{jiang2024mixtral} & Gemini \cite{reid2024gemini} &  Mixtral \cite{jiang2024mixtral} & 24.65 & 27.79 & 11.52 \\
           GPT3.5 \cite{openai2021gpt3.5} & GPT3.5 \cite{openai2021gpt3.5} & GPT3.5 \cite{openai2021gpt3.5} & 25.62 & 29.48 & 12.64 \\
           GPT4 \cite{achiam2023gpt} & GPT4 \cite{achiam2023gpt} & GPT4 \cite{achiam2023gpt} & 25.67 & 29.44 & 12.60 \\
           \hline
           GPT3.5 \cite{openai2021gpt3.5} & Gemini \cite{reid2024gemini}& GPT3.5 \cite{openai2021gpt3.5}&\textbf{26.04} & \textbf{29.85} & \textbf{13.05} \\\hline  
        \end{tabular}
    \end{adjustbox}
    \vspace{-10pt}
\end{table}

\section{Evaluating LLMs in meta-prompting.} In Tab. \ref{tab:meta-prompting}, we present a detailed evaluation of using different LLMs as an evaluator, an optimizer and a generator within our meta prompting framework. 
The first three rows showcase experiments using different LLMs across all three roles. In contrast, rows four and five employ Mixtral only for both generation and optimisation. Notably, Mixtral performed significantly worse as an optimizer compared to GPT-3.5 and Gemini.
In the sixth and seventh row we tried using the same LLM for all three roles. Compared to other models that utilized two LLMs for evaluation, there was no significant performance improvement when using the same generator. Instead, the use of an additional large model introduced extra costs and time burden.
We found that using the same LLM for both roles was less effective than using two separate LLMs.
%
%
%
Ultimately, we choose GPT-3.5 as the optimizer and Gemini as the evaluator to complete the meta prompting process. 

\section{Limitations}
In this section we discus the limitations of our approach. First, VisMaP leverages meta-prompting to generate pseudo-labels, enabling unsupervised long-video summarization with minimal annotation effort. Yet the predictions of the pre-trained source-domain model influence the quality of these pseudo-labels, which can limit performance when faced with a larger domain gap. Improving robustness under greater domain shifts remains an important direction for future research. Another limitation is that our method is able to produce summaries without incorporating context and other modalities, akin to describing a film watched on mute. Incorporating context and other modalities such as audio and transcripts will improve the quality of the summaries and broaden their usefulness to other tasks, such as search and VQA. Other interesting directions include tunable hierarchical summarisation, so that we can extract summaries at any resolution, from a short clip all the way to the full video. Our LLM-based meta prompting component can also be more broadly investigated and made more general and reusable to other applications. In this sense, ViSMaP is a first step in a journey to develop accessible and robust video understanding techniques.

\end{document}